%% file: main.tex
\documentclass{article} % For LaTeX2e
\usepackage{iclr2024_conference,times}

\usepackage{graphicx}
\usepackage{subcaption}
\input{math_commands.tex}

\usepackage{hyperref}
\usepackage{url}

\title{
Distilling LLMs' Decomposition Abilities into Compact Language Models
}

\author{Denis Tarasov \& Kumar Shridhar \\
ETH Zurich\\
\texttt{\{tarasovd,shridhar.kumar\}@ethz.ch} \\
}

\iclrfinalcopy % Uncomment for camera-ready version, but NOT for submission.
\begin{document}

\maketitle

\begin{abstract}
Large Language Models (LLMs) have demonstrated proficiency in their reasoning abilities, yet their large size presents scalability challenges and limits any further customization. In contrast, compact models offer customized training but often fall short in solving complex reasoning tasks. 
This study focuses on distilling the LLMs' decomposition skills into compact models using offline reinforcement learning. We leverage the advancements in the LLM`s capabilities to provide feedback and generate a specialized task-specific dataset for training compact models. 
The development of an AI-generated dataset and the establishment of baselines constitute the primary contributions of our work, underscoring the potential of compact models in replicating complex problem-solving skills\footnote{Our code and dataset are available at \url{https://github.com/DT6A/GSM8K-AI-SubQ}}.

\end{abstract}

\section{Introduction}
Recent strides in Natural Language Processing (NLP) have brought forth powerful Large Language Models (LLMs) like GPT-4 \citep{openai2023gpt}, Claude 2\footnote{\url{https://www.anthropic.com/index/claude-2}}, or Gemini \citep{team2023gemini}. These models not only excel at straightforward tasks such as summarization and sentiment analysis but, with adept prompting, demonstrate proficiency in handling reasoning tasks that demand mathematical and logical abilities \citep{huang2022towards}. Notably, Chain-of-Thoughts (CoT) prompting \citep{wei2022chain} and its variations \citep{kojima2022large, wang2022self} have proven to be promising and relatively simple techniques for enhancing LLMs' reasoning capabilities.

%Within the realm of complex reasoning, the ability to decompose intricate questions into a set of simpler sub-questions represents a crucial component \citep{shridhar2022automatic, zhou2022least, lyu2023faithful}. However, our current understanding suggests that this facet of reasoning remains understudied, with most existing works concentrating on end-to-end solutions for reasoning and not on the specific part of it. Creating datasets and benchmarks for specific task in Deep Learning is one of the driving forces of the entire field \citep{guss2019minerl, vinyals2019grandmaster, fu2020d4rl, kurenkov2023katakomba}.  In this work our goal primary goal is to fill the gap in the reasoning sub-questioning process by providing dataset and baselines for further research in this direction.

%Compounding the challenge is the computational overhead associated with large model sizes, rendering reasoning tasks computationally expensive and time-consuming in cases when the goal is to tune the model. At the same time usage of CoT-like approaches might be expensive due to the fact that the models with the best reasoning abilities are not freely available. In light of this, distilling different components of the reasoning process into smaller models emerges as a promising avenue for research.

Within the realm of complex reasoning, the ability to decompose intricate questions into a set of simpler sub-questions represents a crucial and understudied component \citep{shridhar2022automatic}. While existing works predominantly focus on end-to-end solutions for reasoning \citep{zhou2022least, lyu2023faithful}, the specific aspect of breaking down complex questions into simpler components has received limited attention. The creation of specialized datasets and benchmarks is integral to advancing the field of Deep Learning \citep{guss2019minerl, vinyals2019grandmaster, fu2020d4rl, kurenkov2023katakomba}. This work addresses the gap in understanding and exploration of the reasoning sub-questioning process by providing a dataset and baselines for further research in this direction.

Compounding the challenge is the computational overhead associated with large model sizes, making reasoning tasks computationally expensive and time-consuming when tuning models. Concurrently, approaches similar to Chain-of-Thoughts (CoT) may incur expenses, given that models with superior reasoning abilities are not available for free. In response, distilling distinct components of the reasoning process into smaller models emerges as a promising avenue for research. Decomposition, particularly in the context of teaching smaller models, proves advantageous due to their cost-effectiveness, reduced computational requirements, and accessibility. 

Reinforcement Learning (RL) has demonstrated remarkable success across various domains with recent success in the NLP domain~\cite {bai2022training, bai2022constitutional, openai2023gpt, team2023gemini}. However, some of the most popular approaches like Reinforcement Learning with Human Feedback (RLHF) \citep{ouyang2022training} demand substantial data, and online approaches require extensive interactions with the environment.  Offline RL \citep{levine2020offline}, an alternative that utilizes data directly, holds potential with limited dataset sizes. This subfield has recently witnessed a surge in development, leading to a growth of diverse approaches \citep{kumar2020conservative, fujimoto2021minimalist, kostrikov2021offline, an2021uncertainty, akimov2022let, yang2022rorl, ghasemipour2022so, nikulin2023anti}. Numerous successful applications of offline RL exist in fields like robotics~\citep{smithWalkParkLearning2022, kumar2021a}, autonomous driving \citep{Diehl2021UMBRELLAUM}, recommendation systems\citep{Chen2022OffPolicyAF} and even drug-design \citep{tarasov2023offline}. For the NLP domain, recent studies have intriguingly revealed that AI feedback closely resembles human feedback \citep{lee2023rlaif}, and language models can be fine-tuned using their own generated feedback \citep{bai2022constitutional}.  

In this preliminary work, we combine the advantages of sub-questioning with offline RL with feedback for the task of mathematical reasoning and distill the subquestion decomposition abilities in smaller models. Our work proposes the following: 1) an AI-generated benchmark where math questions are broken down into simpler sub-questions based on the GSM8K \citep{cobbe2021training} dataset, 2) train smaller language models for the same task using fine-tuning and offline RL techniques to provide baselines for the task, and 3) explore the potential benefits of using AI-generated feedback on its own responses in enhancing model performance. Our experimental results reveal a big gap between ChatGPT's reasoning abilities and what can be achieved with smaller models and existing algorithmical approaches.

\section{Related Work}

\subsection{LM Distillation}
Distillation has emerged as a pivotal technique in mitigating the computational challenges associated with LLMs while retaining their valuable knowledge. Notably, \citet{hinton2015distilling} introduced knowledge distillation as a means to transfer the knowledge from a complex model to a simpler one, enabling the latter to approximate the former's performance. In the context of language models, \citet{sanh2019distilbert} successfully distilled BERT, a prominent LLM, into a more compact version named DistilBERT, maintaining competitive performance with significantly fewer parameters. Recently, a lot of work has successfully distilled the reasoning capabilities into smaller models \cite{Yuan2023ScalingRO, magister2023teaching, shridhar2023distilling, Hsieh2023DistillingSO}

\subsection{LLMs Reasoning}
The exploration of reasoning capabilities in Large Language Models (LLMs) has been a focal point in recent NLP research. \citet{wei2022chain} introduced Chain-of-Thoughts (CoT), a method compelling LLMs to approach problems in a step-by-step manner rather than providing direct answers. Noteworthy is their revelation that reasoning abilities manifest prominently in larger LM sizes. This technique involves prompting the model with step-by-step problem-solving and furnishing multiple examples for guidance through few-shot prompting. Building on this, \citet{kojima2022large} demonstrated that reasoning abilities may emerge even without examples, utilizing zero-shot prompting. Further refinement in reasoning quality was achieved by \citet{wang2022self}, who showed that applying CoT multiple times and picking the most frequent answer boosts resulting performance.

\subsection{Sub-Questioning}
The paradigm of splitting reasoning into sub-question generation and sub-question answering has proven beneficial in solving reasoning problems \citep{shridhar2022automatic}. Notably, \citet{shridhar2023distilling} took a step further by distilling LLMs reasoning abilities into substantially smaller models through finetuning, laying the groundwork for our current work. It is essential to note that both of these works primarily address the entire reasoning process rather than dissecting its individual components.
In a parallel thread of research, \citet{zhou2022least} demonstrated that decomposing reasoning problems into sub-problems results in improved LLM performance. Their findings indicate that LLMs can execute this split step-by-step, outperforming CoT-based approaches.

Extending beyond mere decomposition, \citet{juneja2023small} introduced a finetuning approach where a ``helper" LM poses questions to a ``solver" LM, guiding it toward the correct solution based on the problem and the sequence of prior interactions. Despite showcasing the efficacy of this approach, its reliance on an online RL paradigm and the use of large 13 or 33 billion parameters LLaMA models \citep{touvron2023llama} poses practical challenges. The inherent computational demands and potential instability of online RL, compounded by the necessity for significant interactions and potential cost implications, underscore the complexities associated with this method.

\section{Preliminaries}
\subsection{Language Modeling}
Our study leverages the preeminence of attention-based Transformer architectures, as introduced by \citet{vaswani2017attention}. In the realm of autoregressive models, we turn our attention to the widely adopted GPT-2 architecture \citep{radford2019language}, specifically selecting models of various sizes to tailor our experiments. The chosen finetuning strategies center around the application of standard cross-entropy loss, optimizing the models for accurate next-token prediction. For every approach we provide mathematical problem as a prefix and expect model to generate the sub-questions required to solve the problem.

\subsection{Offline RL task formulation}
Given the recent advancements in offline RL, we incorporate a baseline from this field. Building upon the formulation proposed by \citet{snell2022offline}, we cast the text generation problem as a token-level Partially Observable Markov Decision Process (POMDP). In this framework, the agent's observations correspond to prefixes of tokens, and the agent's action pertains to the selection of the next token to be generated. 

\section{GSM8K-AI-SubQ Dataset}
In the pursuit of advancing research in reasoning within Language Models (LMs), we introduce the GSM8K-AI-SubQ dataset, uniquely designed to emphasize sub-questioning and leverage AI-generated feedback for these sub-questions. Our inspiration comes from the works of \citet{bai2022constitutional} and \citet{lee2023rlaif}. The first work provide compelling evidence that such feedback can serve as a valuable signal for fine-tuning models. The second work shows  that AI-generated data exhibits comparable quality to human-annotated data in the context of preferences. We hope that our dataset will provide any insights whether similar properties hold for the reasoning.

While acknowledging that AI models may not offer the ideal source of reasoning data, we posit the substantial benefits of such a dataset. Firstly, it serves as a valuable resource for distilling reasoning abilities into smaller models, aligning with the growing trend of efficiency in language model architectures. Importantly, even if the data exhibits suboptimal characteristics, we draw inspiration from the success of offline RL in other fields, where it has demonstrated an ability to outperform the policy responsible for dataset collection. This resilience to suboptimal data quality mitigates concerns and underscores the dataset's potential impact on advancing reasoning capabilities. Lastly, the expedited and cost-effective nature of this data acquisition method democratizes its accessibility, allowing researchers to extend our dataset if required.

For the sake of completeness, we furnish LLM responses for the generated sub-questions, although our primary focus in this work centers on the sub-questioning. This dataset not only facilitates advancements in LM reasoning but also lays the groundwork for innovative approaches in the broader landscape of language model research.

\subsection{Dataset Collection Process}
In our dataset collection process, we leverage the capabilities of ChatGPT, specifically utilizing the \texttt{gpt-3.5-turbo-0613}\footnote{The most recent  \texttt{gpt-3.5-turbo} version} version. This version is chosen due to its lenient restrictions on the number of queries and its cost-effectiveness, offering responses of commendable quality when benchmarked against similar LLMs \citep{tunstall2023zephyr}. Interactions with ChatGPT are facilitated through the OpenAI API\footnote{\url{https://platform.openai.com/docs/overview}}, employing a temperature setting of 0.7 to ensure a diverse range of responses.

For the specific task at hand, we curate a dataset of mathematical problems extracted from the GSM8K dataset \citep{cobbe2021training}. This dataset comprises a diverse array of grade school math word problems meticulously crafted by human problem writers. While designed to be solvable by a middle school student, these problems present a noteworthy challenge for LMs. Maintaining the integrity of the original train/test split, our dataset consists of 7473 training examples and 1319 testing examples.

\textbf{Sub-questions Generation.} The initial phase of our data generation involves creation of set of sub-questions for each problem within the GSM8K dataset.  We prompt ChatGPT to decompose the given problem into distinct sub-problems and provide corresponding sub-questions, aiming to elucidate the problem-solving process for others. To guide the model and ensure a consistent output format, we furnish two examples of problems and their corresponding sub-questions as demonstration inputs. Each request is treated as an independent dialogue to eliminate potential interference. We keep a 2-shot prompting strategy and preserving independence for all subsequent queries. A sample input and output for this sub-question generation process are illustrated in \autoref{tab:prompt_qg}. To enhance the dataset's size and diversity, we repeat the inquiry for each problem three times, resulting in 22,419 training samples. While the repetition could be further increased for creating a larger dataset, we acknowledge budget constraints as a limiting factor in this study.

\textbf{Answers Generation.} In the subsequent phase of our data collection, we focused on generating responses to the previously obtained sub-questions. Tasking ChatGPT solely with the responsibility of providing answers to these queries and obtaining a final answer for the original problem. An illustrative example of this prompt-response interaction is presented in \autoref{tab:prompt_qa}. The generated final answers were then employed as labels to categorize each set of sub-questions. Sub-questions associated with an original problem solution were identified as "good", while those failing to lead to a resolution were categorized as "bad". 

\textbf{Feedback Generation.} Concluding our dataset collection, we engage in the pivotal task of soliciting feedback from ChatGPT on its generated sub-questions. The objective is to introduce a nuanced signal at the individual question level, mitigating sparsity concerns associated with relying solely on the correctness of final answers. This aspect assumes significance, especially in the context of potential RL applications. Drawing inspiration from the effective sub-questioning strategy proposed by \citet{shridhar2022automatic}, we task ChatGPT with determining the usefulness of each sub-question in the problem-solving process. Refer to \autoref{tab:prompt_feedback} for an exemplary interaction. To account for potential inconsistencies in ChatGPT's feedback, we query the model three times for each set of sub-questions. This repetition strategy aligns with findings in recent studies \citep{wang2022self, lee2023rlaif}, showing that leveraging multiple responses often results in more useful data. It's worth noting that the repetition factor can be adjusted for further data refinement. Subsequently, scores for each sub-question are computed by evaluating the fraction of responses deeming it useful, establishing a metric for assessing the efficacy of individual sub-questions.

\textbf{Dataset Collection Costs.} The compilation of our training set demanded approximately 12 hours of real-time investment and incurred a cost of approximately \$100 for utilizing the \texttt{gpt-3.5-turbo} version. Notably, feedback generation constituted around 70\% of the total costs, primarily attributed to the necessity of acquiring feedback multiple times for each set of questions.

\subsection{Dataset Analysis}

\begin{table}[h]
  \centering
  \begin{tabular}{l|cccc}
    \hline
    \textbf{Metric} & \textbf{0 correct} & \textbf{1 correct} & \textbf{2 correct} &  \textbf{3 correct} \\
    \hline
    Number of problems & 1343 (18\%) & 866 (11\%) & 1139 (15\%) & 4052 (54\%) \\
    \hline
    Mean problem length & 269.4 $\pm$ 106.7 & 252.1 $\pm$ 100.3 & 240.2 $\pm$ 94.7 &  217.5 $\pm$ 82.9 \\
    \hline
    Median problem length & 250 & 235.5 & 225 & 201.5 \\
    \hline
  \end{tabular}
  \caption{Train set statistics for problems splitted by the number of sub-question sets which lead to the correct answer. "Problem length" denotes number of characters.}
  \label{tab:train_stats}
\end{table}

In this section, we present an analysis of the collected training data, offering valuable statistics and insights.

\begin{figure}[ht]
\centering
\captionsetup{justification=centering}
     \centering
     \begin{subfigure}[b]{0.49\textwidth}
         \centering
         \includegraphics[width=\textwidth]{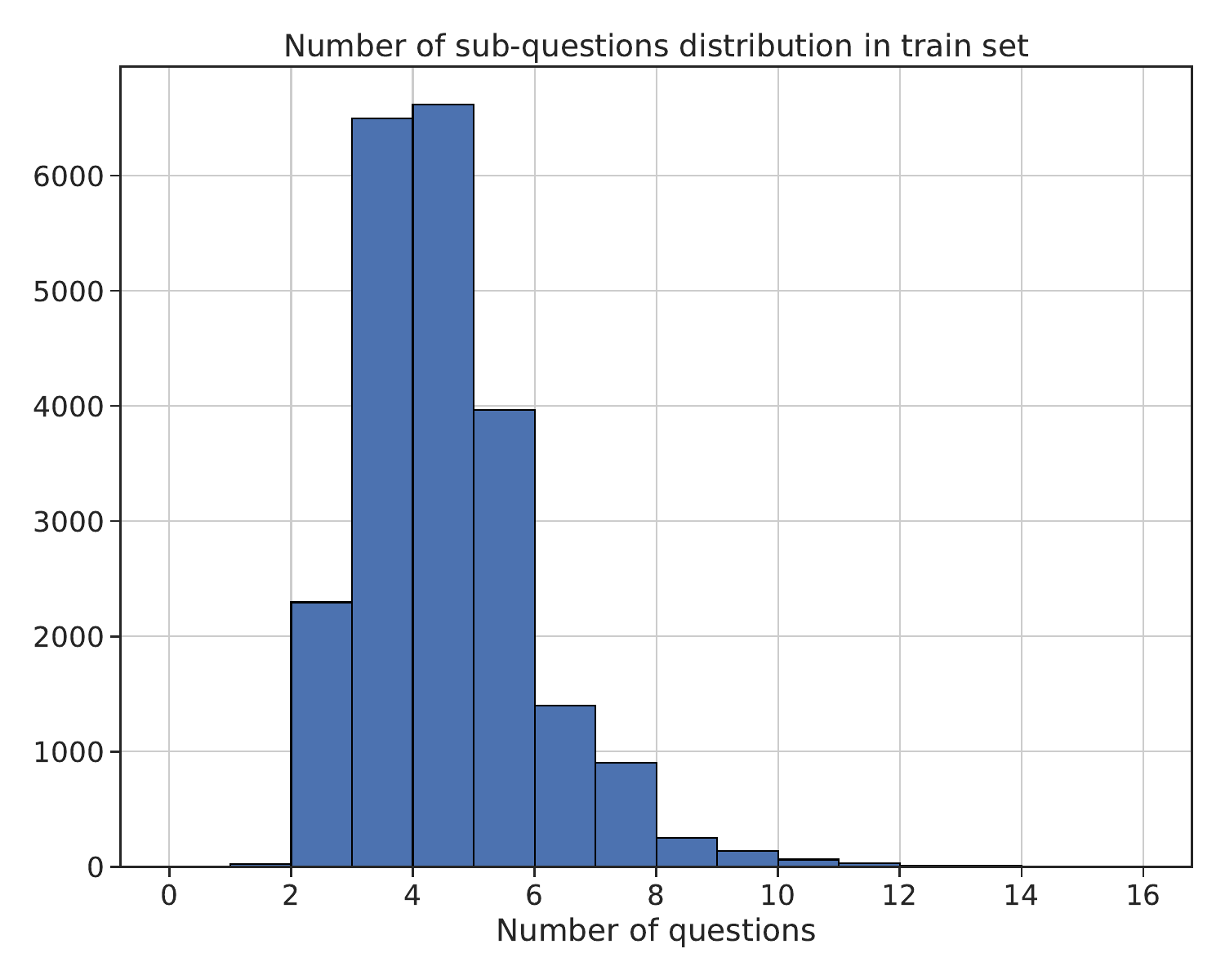}
         \caption{}
         \label{fig:train_all_q_distr}
     \end{subfigure}
     \hfill
     \begin{subfigure}[b]{0.49\textwidth}
         \centering
         \includegraphics[width=\textwidth]{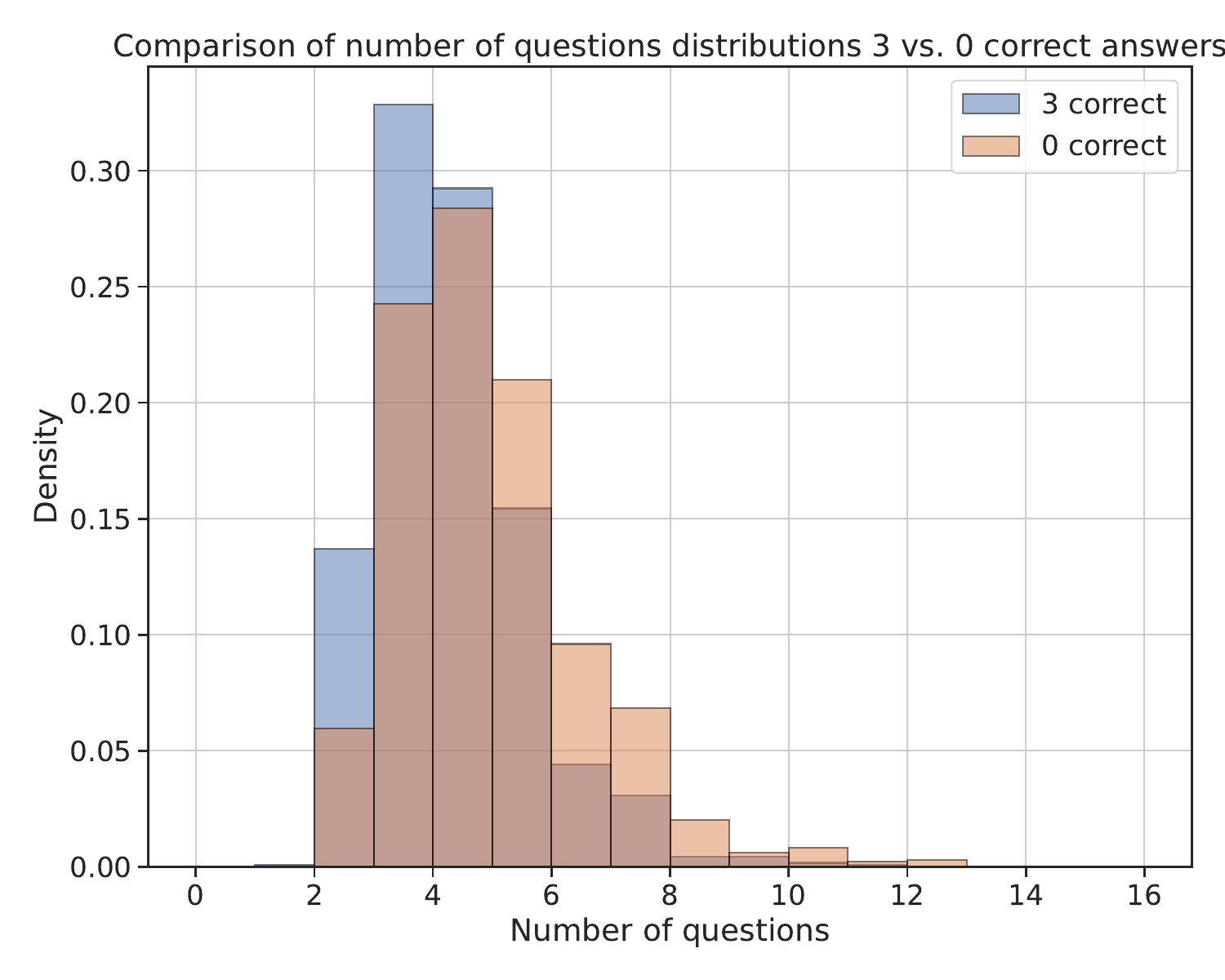}
         \caption{}
     \end{subfigure}
     \begin{subfigure}[b]{0.49\textwidth}
         \centering
         \includegraphics[width=\textwidth]{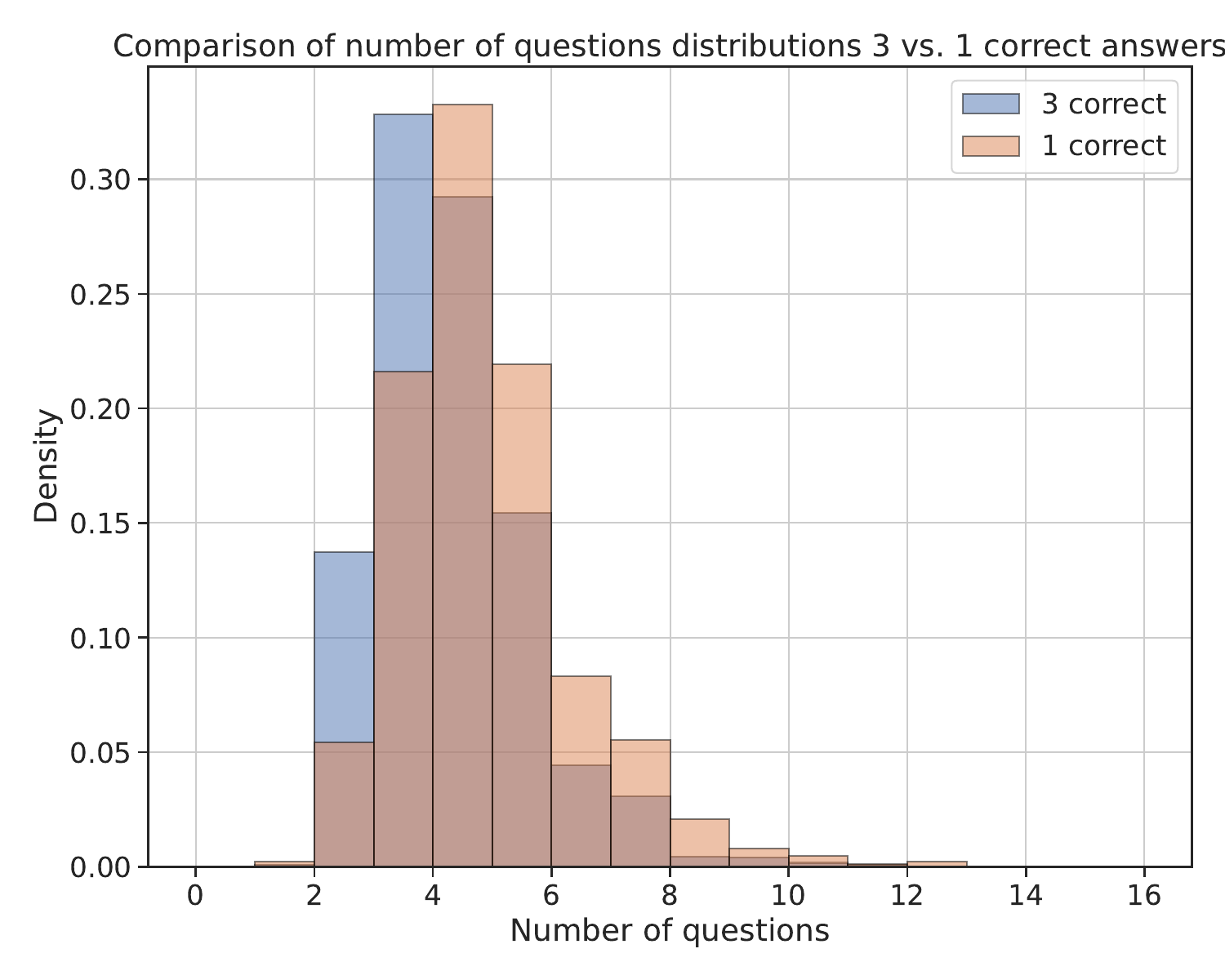}
         \caption{}
     \end{subfigure}
     \begin{subfigure}[b]{0.49\textwidth}
         \centering
         \includegraphics[width=\textwidth]{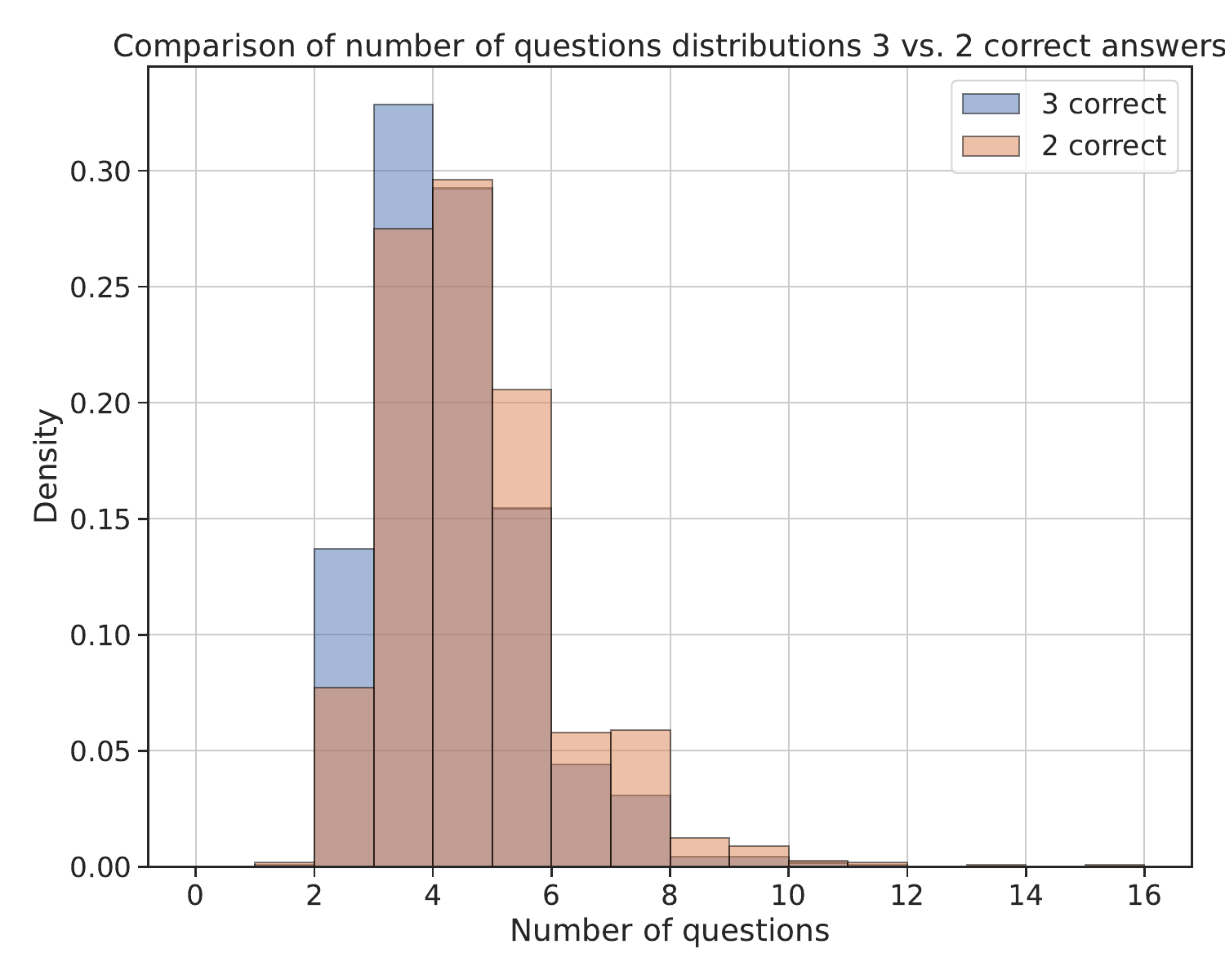}
         \caption{}
     \end{subfigure}
    \caption{Number of questions distributions in train set. (a) Distribution in the entire train set with mean 4.0 $\pm$ 1.5 and median 4, (b, c, d) Comparisons of distributions in number of questions between problems with 3 vs. 0, 1, 2 sets of sub-questions that lead to the correct final answer.}
        \label{fig:train_q_dist}
\end{figure}

The distribution of sub-question set sizes is visualized in \autoref{fig:train_all_q_distr}. While the majority of sets comprise 2 to 6 questions, some outliers exist, featuring either a single question or more than 6. Notably, we opt to retain these outliers in our dataset for a more diverse representation.

Our analysis extends to evaluating the efficacy of sub-question sets in leading to the correct solution for each problem. As outlined in \autoref{tab:train_stats}, approximately 54\% of the sets consistently resulted in a correct solution across all three attempts, while only around 18\% failed to yield a correct solution in any instance. Moreover, longer problems appear to pose increased difficulty, aligning intuitively with expectations. Calculating average accuracy by treating each sub-question set as an independent problem, the overall accuracy for the training set stands at {0.68}.

To delve deeper into the relationship between problem complexity and sub-question count, we compare distributions for 0, 1, and 2 out of 3 correct answers versus 3 out of 3 correct answers in \autoref{fig:train_q_dist} b, c, d. The analysis reveals a discernible trend: easier problems tend to be associated with a smaller number of sub-questions.

Continuing our exploration, we investigate whether ChatGPT's feedback on its own sub-questions holds intrinsic value and diverges from random noise.

\begin{figure}[ht]
\centering
\captionsetup{justification=centering}
     \centering
     \begin{subfigure}[b]{0.32\textwidth}
         \centering
         \includegraphics[width=\textwidth]{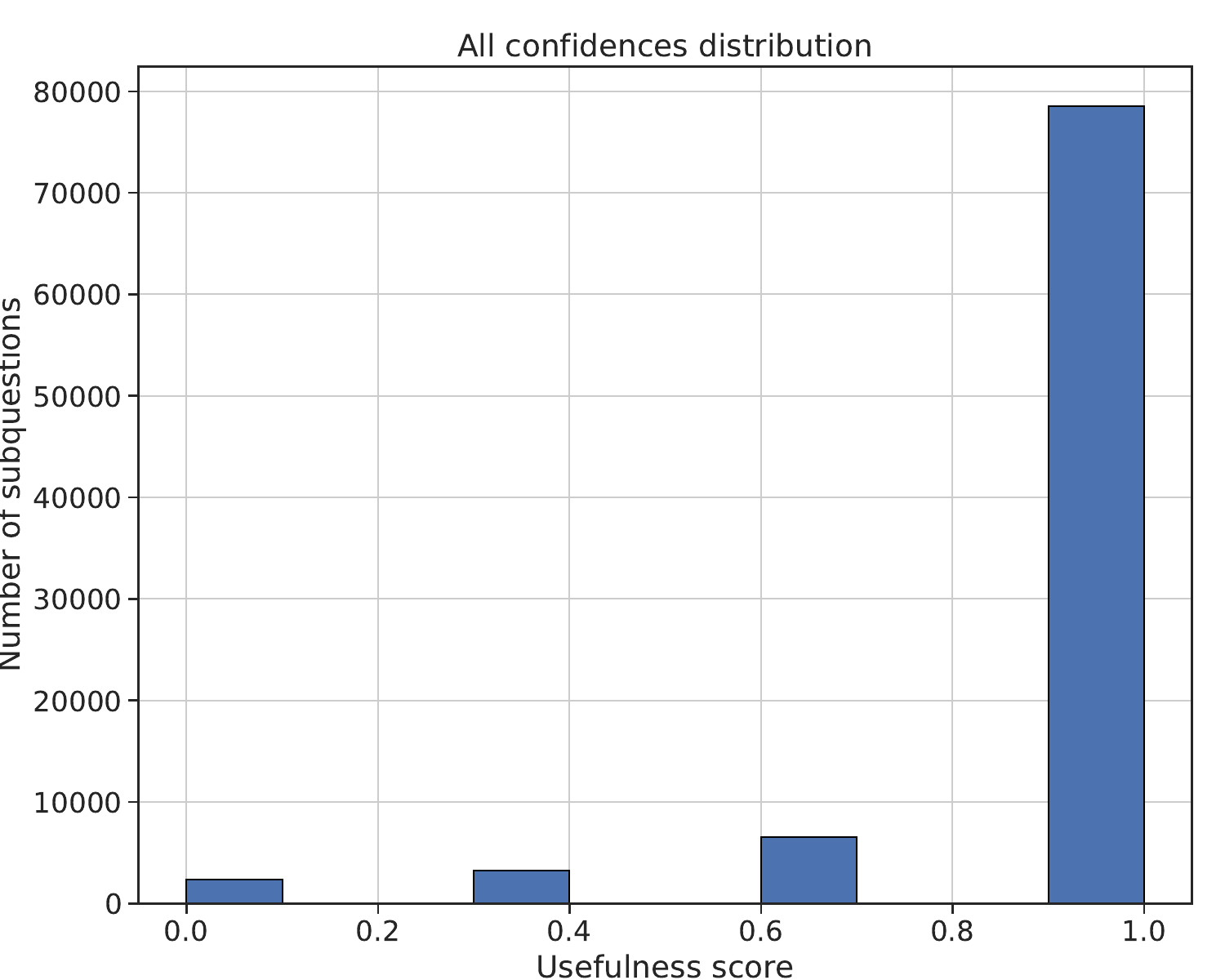}
         \caption{}
         \label{fig:all_feedback_distr}
     \end{subfigure}
     \hfill
     \begin{subfigure}[b]{0.32\textwidth}
         \centering
         \includegraphics[width=\textwidth]{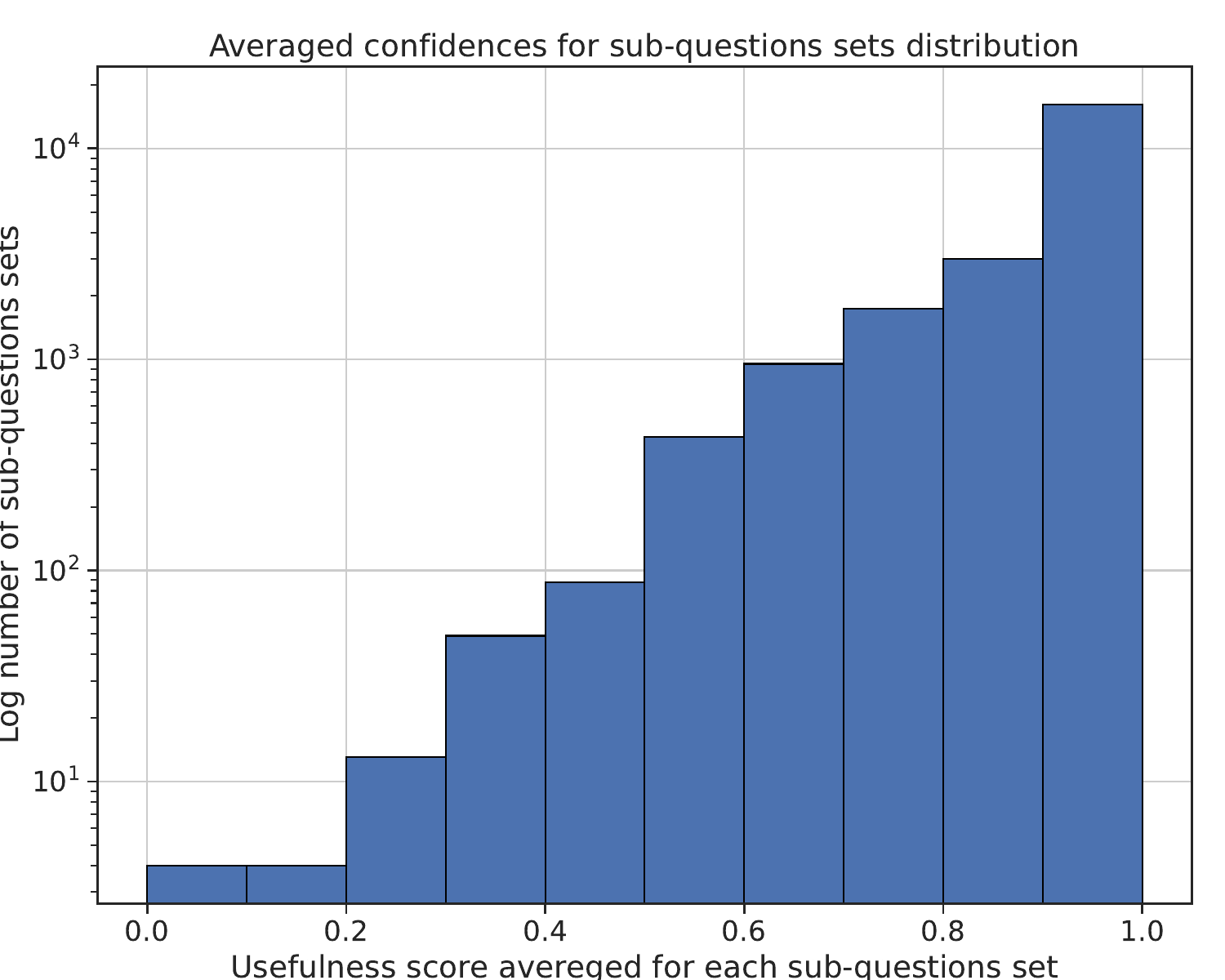}
         \caption{}
         \label{fig:problems_feedback_distr}
     \end{subfigure}
     \hfill
     \begin{subfigure}[b]{0.32\textwidth}
         \centering
         \includegraphics[width=\textwidth]{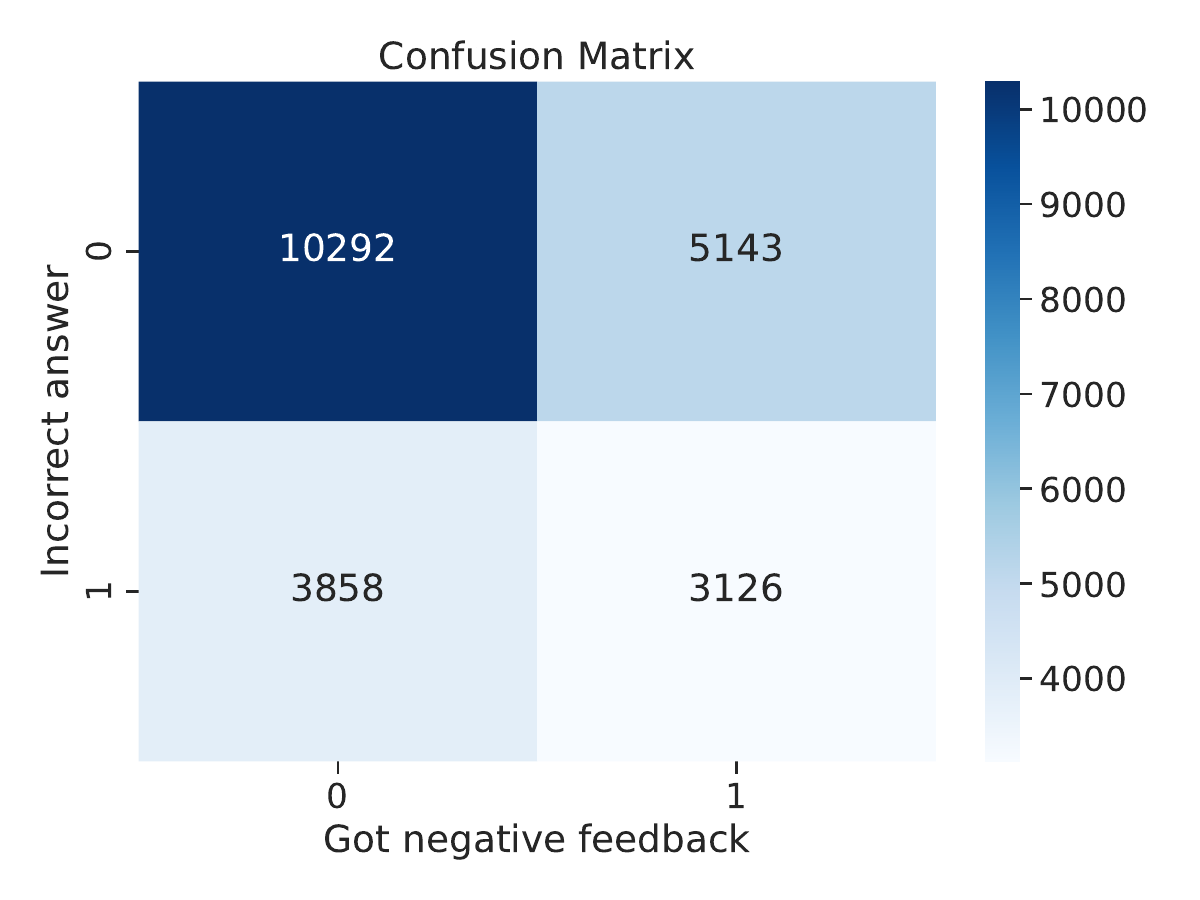}
         \caption{}
         \label{fig:feedback_cm}
     \end{subfigure}
    \caption{(a) distribution of the usefulness feedback over individual sub-questions, (b) distribution of the usefulness feedback averaged over sets of sub-questions, (c) confusion matrix between the presence of negative feedback in the set of sub-questions and correctness of the final answer based on corresponding sub-questions.}
    \label{fig:feedback_distrs}
\end{figure}

The distribution of usefulness scores for each sub-question, depicted in \autoref{fig:all_feedback_distr}, indicates that approximately 90\% of sub-questions received consistent markings as useful across all three iterations. However, to gain a more nuanced perspective, we average the usefulness scores for each sub-question set and showcase the distribution of averaged confidences in \autoref{fig:problems_feedback_distr}.

Upon examining the fraction of problems with at least one negative response, we find that 36\% of problems fall into this category, aligning with the 54\% that were correctly solved 3 out of 3 times. Recognizing that a single misstep in the reasoning process can lead to an incorrect answer, we employ the presence of negative feedback as an indicator for predicting an incorrect solution. However, this indicator comes with limitations, such as the potential for false negatives arising from the model providing the correct final answer despite a flawed reasoning step, and the occasional presence of randomness during feedback collection. The confusion matrix in \autoref{fig:feedback_cm} outlines this setup, showcasing differences from a random scenario. Notably, when negative feedback is absent, 72\% of problems were actually solved correctly, compared to 62\% when negative feedback is present.

To quantitatively assess the performance of this heuristic, we compute the ROC AUC score, treating the average sub-question set usefulness as a probability and designating an incorrect answer as the "target label". The resulting ROC AUC score of 0.56 indicates a departure from randomness (0.5). Further, the Pearson correlation coefficient, calculated at -0.09 with a p-value of $10^{-46}$, suggests that a higher usefulness score corresponds to a decreased likelihood of encountering an incorrect answer. Despite the simplicity of this heuristic, it provides informative cues on the efficacy of the sub-question set, indicating its potential utility.

\section{Baselines}
In our exploration of the constructed dataset, we conduct a series of experiments utilizing both supervised techniques and an offline RL approach. Our experiments involve the deployment of pretrained small and medium versions of GPT-2 \citep{radford2019language} as well as pretrained DistilGPT \citep{sanh2019distilbert} to gauge performance dynamics across various models sizes. The GPT-2 medium model, with 345 million parameters, represents the largest model in our experiments. This choice is deliberate, as larger models endowed with reasoning abilities often exceed 7 billion parameters, rendering their usage complex for many researchers. Unfortunately, due to computational resource limitations, our experiments did not extend to GPT-2 large and XL models.

\subsection{Evaluation Protocol}  
To facilitate a robust comparison of different approaches, we employ the same version of ChatGPT utilized during data collection. Sub-questions generated by each approach are presented to ChatGPT using the same prompt from \autoref{tab:prompt_qa}, and we measure the accuracy of final answers as our evaluation metric. Standardizing the evaluation environment, we set the temperature to 0 and fix the random seed to ensure reproducibility and eliminate randomness during the evaluation process. This rigorous evaluation protocol ensures a fair and consistent assessment of the performance across various approaches. One evaluation of test set costed approximately 1.5\$ with \texttt{gpt-3.5-turbo}. 

We additionally use the same protocol to evaluate results using open-source models Mistral 7B \citep{jiang2023mistral}, LLaMA 7B and LLaMA 13B \citep{touvron2023llama}.

\subsection{Applied Approaches}
In this subsection, we provide a brief overview of each applied approach. For a more details, refer to \autoref{app:details}.

\textbf{Behavioral Cloning.} Behavioral Cloning (BC) is a fundamental and robust approach commonly applied to datasets involving decision-making tasks. In the context of NLP, it translates to fine-tuning a language model to replicate a specific behavior or policy observed in the dataset. To select the best model, we employ a small held-out fraction (1\%) of the training data. Following the methodology inspired by \citet{shridhar2022automatic}, we use the BLEU score \citep{papineni2002bleu} calculated between the generated questions and those produced by ChatGPT. Given that the primary goal of BC is to replicate the original policy's behavior, BLEU serves as a suitable metric, indicating the similarity between two texts. Our empirical observation show that BLEU correlates with the final performance, making it a reasonable choice for model evaluation in the context of BC. For all subsequent approaches, the best BC model serves as the initialization for the LM.

\textbf{Filtered Behavioral Cloning.} Filtered BC \citep{chen2021decision} introduces a modification of BC by considering only a fraction of the best trajectories in the dataset. This approach proves particularly effective when a substantial number of high-quality examples are at disposal. In the context of our task, we exclusively retain samples corresponding to sub-question sets that result in the correct solution. The model selection process remains consistent with the standard BC approach.

\textbf{Implicit Language Q-Learning.} Implicit Language Q-Learning (ILQL) \citep{snell2022offline} represents an adaptation of the offline RL approach known as IQL \citep{kostrikov2021offline} to NLP tasks. The core idea behind ILQL involves training additional Value (V) and Q-function heads with IQL objectives. These additional functions are then employed to reweight the original LM outputs using the advantage value, which is the difference between V and Q values.

The selection of ILQL is motivated by the effectiveness of IQL as one of the strongest offline RL approaches in diverse domains \citep{tarasov2022corl}. Given the limited adaptation of offline RL approaches to NLP problems, ILQL emerges as the state-of-the-art choice. Given that IQL optimizes for rewards, which may not inherently correlate with the dataset policy, selecting the best model becomes challenging. In the absence of a clear best model selection criterion, we have tried to use the same criterion as we did for BC and the common offline RL practice of taking the latest checkpoint after training. The first method produced slightly better results on average.

Two versions of ILQL are tested in our experiments: ILQL-full utilizing all available rewards from the dataset and ILQL-sparse employing only answer correctness as rewards.

\subsection{Experimental Results}

The experimental results, summarized in \autoref{tab:eval}, offer insights into the performance of the proposed approaches averaged over various answering LLMs. For a detailed breakdown based on specific answering models, refer to \autoref{app:scores}.

\begin{table}[h]
  \centering
  \begin{tabular}{l|cccc}
    \hline
    \textbf{Algorithm} & \textbf{DistillGPT} & \textbf{GPT-2 small}& \textbf{GPT-2 medium} & \textbf{Average} \\
    \hline
    BC & 0.255 & 0.284 & 0.310 & 0.283 \\
    \hline
    Filtered BC & \textbf{0.260} & \textbf{0.293} & \textbf{0.319} & \textbf{0.291}\\
    \hline
    ILQL-sparse & 0.249 & 0.281 & 0.310 & 0.280\\
    \hline
    ILQL-full & 0.253 & 0.277 & 0.309 & 0.280 \\
    \hline
    \hline
    ChatGPT & N/A & N/A & N/A & 0.429\\
    \hline
  \end{tabular}
  \caption{Accuracy score of the final answer avereged over different models which were used for sub-questions answering. Best scores are highlighted with \textbf{bold}.}
  \label{tab:eval}
\end{table}

It is evident that all tested approaches fall short when compared to the ChatGPT, indicating substantial room for improvement. Sub-questions generation abilities also improve with the size of the backbone model which alligns with previous researches.

Filtered BC demonstrates improved performance over standard BC in most scenarios, consistent with expectations when sufficient amount of high-quality demonstrations are present in the dataset. However, this trend is not universal, particularly when LLaMAs serve for question answering and GPT-2 medium is a backbone model for sub-question generation. Interestingly, Filtered BC outperforms BC only when GPT-2 medium is used alongside Mistral for answering.

The comparison between ILQL-sparse and ILQL-full does not reveal a consistent advantage for either method. In most of the cases, both ILQL variants underperform Filtered BC, and even falling behind standard BC in half of the cases. However, the superiority of the Filtered BC over offline RL approaches in NLP was also recently demonstrated by \citet{gulcehre2023reinforced}. Our finding should solve as additional motivation for the development of offline RL algorithms for NLP.

\section{Limitations and Future Work.}
Our work serves as a foundational exploration, opening avenues for various future directions.

\textbf{Development of Offline RL Approaches:} A pivotal area for future exploration involves advancing offline RL or other suitable methodologies for distilling reasoning abilities from static datasets. This extension could contribute to more effective utilization of language models in reasoning tasks.

\textbf{Creation of a Larger Benchmark:} Expanding our methodology, future work could focus on generating a more extensive benchmark as it requires only the access to ground truth-answers in the datset which usualy holds. This benchmark might incorporate a diverse set of reasoning datasets, such as MATH \citep{hendrycks2021measuring} or AQuA \citep{ling2017program}, providing a broader assessment of reasoning capabilities.

\textbf{Concentration on Sub-Question Answering:} Delving deeper into the sub-question answering aspect of the reasoning process presents a promising direction. While our dataset includes ChatGPT responses for sub-questions, their scoring and utilization remain unexplored. Future studies could investigate this component to enhance understanding and performance.

\textbf{Utilization of Open-Source Models:} Exploring the application of open-source models, such as LLaMA, for sub-question generation emerges as a cost-effective alternative. Accessible without financial constraints, these models present an opportunity for researchers to delve into sub-question generation without monetary limitations. We were not able to run such kind of experiments ourselves due to the computational limitations.

\section{Conclusion}
This work introduces a novel AI-generated benchmark tailored for evaluating sub-questioning in reasoning tasks. We employ diverse offline learning approaches, varying model sizes for baselines, and assess the performance using different LLMs. Our experiments aim to shed light on the challenges and potential avenues for enhancing reasoning capabilities.

The outcomes reveal a significant performance gap between the best-performing approach and ChatGPT. The underwhelming performance of the offline RL approach underscores the need for further advancements in this domain, presenting an opportunity for future research to explore and refine these methodologies.

By providing this benchmark, we aspire to catalyze research endeavors in the realm of sub-questioning. We anticipate that the dataset curated in this work will serve as a foundational resource for assessing the reasoning capabilities of emerging offline RL approaches in the field of NLP. 

\section{Acknowledgements}
We thank Mrinmaya Sachan for his supervision of our work and Elliott Ash for providing us with computational resources.

\bibliography{iclr2024_conference}
\bibliographystyle{iclr2024_conference}

\clearpage

\appendix

\section{Experimental Details}
\label{app:details}
We run all of our experiments using single V100 GPUs. The training time never exceeded 5 days.

Hyperparaters are kept the same across different model sizes.
\subsection{Hyperparameters}
We conducted hyperparameters search only through $\{0.5, 0.7, 0.9\}$ values for IQL $\tau$ and $\{0.5, 1.0, 3.0\}$ values for ILQL $\beta$ parameter using GPT-2 small with full reward. Also, following \citet{tarasov2023revisiting} we have increased discount factor value from default 0.99 to 0.999 which improved ILQL performance.

\begin{table}[!ht]
\centering
\begin{tabular}{cl|l}
\hline
& Hyperparameter & Value \\
\hline
 & Batch size   & 32        \\
  & Optimizer & Adam \\
  & Learning rate   & 1e-4       \\
  & Gradient steps & 10000 \\
\hline
\end{tabular}
\caption{BC hyperparameters.}\label{tab:hyp_bc}
\end{table}

\begin{table}[!ht]
\centering
\begin{tabular}{cl|l}
\hline
& Hyperparameter & Value \\
\hline
 & Batch size   & 32        \\
  & Optimizer & Adam \\
  & Learning rate   & 1e-4       \\
  & Gradient steps & 7500 \\
\hline
\end{tabular}
\caption{Filtered BC hyperparameters.}\label{tab:hyp_fbc}
\end{table}

\begin{table}[!ht]
\centering
\begin{tabular}{cl|l}
\hline
& Hyperparameter & Value \\
\hline
 & Batch size   & 32        \\
  & Optimizer & Adam \\
  & Learning rate   & 1e-4       \\
  & Gradient steps & 25000 \\
  & Discount factor      & 0.999 \\
  & Target update rate   & 5e-3 \\
  & IQL $\tau$            & 0.9 \\
  & ILQL $\beta$            & 1.0 \\
  & V loss weight         & 1.0 \\
  & Q loss weight         & 1.0 \\
  & CQL loss weight       & 0.01 \\
\hline
\end{tabular}
\caption{ILQL hyperparameters.}\label{tab:hyp_ilql}
\end{table}

\section{Full Tabular Scores}
\label{app:scores}
\begin{table}[h]
  \centering
  \begin{tabular}{l|cccc}
    \hline
    \textbf{Algorithm} & \textbf{DistillGPT} & \textbf{GPT-2 small}& \textbf{GPT-2 medium} & \textbf{Average} \\
    \hline
    BC & 0.476 & 0.508 & 0.538 & 0.507 \\
    \hline
    Filtered BC & \textbf{0.493} & \textbf{0.527} & \textbf{0.576} & \textbf{0.532}\\
    \hline
    ILQL-sparse & 0.471 & 0.518 & 0.541 & 0.510 \\
    \hline
    ILQL-full & 0.484 & 0.504 & 0.540 & 0.509\\
    \hline
    \hline
    ChatGPT & N/A & N/A & N/A & 0.682\\
    \hline
  \end{tabular}
  \caption{Accuracy score of the final answer using ChatGPT for sub-questions answering. Best scores are highlighted with \textbf{bold}.}
  \label{tab:eval_chat}
\end{table}

\begin{table}[h]
  \centering
  \begin{tabular}{l|cccc}
    \hline
    \textbf{Algorithm} & \textbf{DistillGPT} & \textbf{GPT-2 small}& \textbf{GPT-2 medium} & \textbf{Average} \\
    \hline
    BC & 0.118 &  0.154 & \textbf{0.164} & 0.145\\
    \hline
    Filtered BC & \textbf{0.125} & \textbf{0.159} & 0.162 & \textbf{0.149}\\
    \hline
    ILQL-sparse & \textbf{0.125} & 0.138 & 0.162 & 0.142 \\
    \hline
    ILQL-full & 0.114 & 0.144 & 0.163 & 0.140 \\
    \hline
    \hline
    ChatGPT & N/A & N/A & N/A & 0.234 \\
    \hline
  \end{tabular}
  \caption{Accuracy score of the final answer using LLaMA 7B for sub-questions answering. Best scores are highlighted with \textbf{bold}.}
  \label{tab:eval_llama_7b}
\end{table}

\begin{table}[h]
  \centering
  \begin{tabular}{l|cccc}
    \hline
    \textbf{Algorithm} & \textbf{DistillGPT} & \textbf{GPT-2 small}& \textbf{GPT-2 medium} & \textbf{Average} \\
    \hline
    BC & 0.184 & 0.212 & 0.247 & 0.214\\
    \hline
    Filtered BC & \textbf{0.194} & \textbf{0.230} & 0.245 & \textbf{0.223} \\
    \hline
    ILQL-sparse & 0.180 & 0.207 & {0.250} & 0.212\\
    \hline
    ILQL-full & 0.182 & 0.210 & \textbf{0.252} & 0.215\\
    \hline
    \hline
    ChatGPT & N/A & N/A & N/A & 0.353\\
    \hline
  \end{tabular}
  \caption{Accuracy score of the final answer using LLaMA 13B for sub-questions answering. Best scores are highlighted with \textbf{bold}.}
  \label{tab:eval_llama_13b}
\end{table}

\begin{table}[h]
  \centering
  \begin{tabular}{l|cccc}
    \hline
    \textbf{Algorithm} & \textbf{DistillGPT} & \textbf{GPT-2 small}& \textbf{GPT-2 medium} & \textbf{Average} \\
    \hline
    BC & \textbf{0.240} & \textbf{0.264} & 0.290 & \textbf{0.265}\\
    \hline
    Filtered BC & 0.228 & 0.256 & \textbf{0.293} & 0.259 \\
    \hline
    ILQL-sparse & 0.219 & 0.261 & 0.288 & 0.256 \\
    \hline
    ILQL-full & 0.231 & 0.252 & 0.280 & 0.254 \\
    \hline
    \hline
    ChatGPT & N/A & N/A & N/A & 0.446\\
    \hline
  \end{tabular}
  \caption{Accuracy score of the final answer using Mistral for sub-questions answering. Best scores are highlighted with \textbf{bold}.}
  \label{tab:eval_mistral}
\end{table}

\clearpage
\section{ChatGPT Prompts}
\label{app:prompts}

\begin{table}[h]
  \centering
  \begin{tabular}{p{15cm}}
    \hline
    \textbf{Input}\\
    \hline
    You are given mathematical problems marked with "Problem". Your task is to split it into smaller sub-problems and formulate  them as sub-questions which will be answered by someone else who's objective is to solve the original problem. Questions must not contain the answers for the previous questions in them. Do not ask questions where the answer is already given in the problem. For each problem come up with the sequence of sub-questions and output each of them on separate line which starts with letter Q followed by the number of question. Do not output anything else.
\\\\
Problem: A robe takes 2 bolts of blue fiber and half that much white fiber.  How many bolts in total does it take?
\\\\
Q1: How many bolts of white fiber does it take? \\
Q2: How many bolts in total does it take? 
\\\\
Problem: Betty is saving money for a new wallet which costs \$100. Betty has only half of the money she needs. Her parents decided to give her \$15 for that purpose, and her grandparents twice as much as her parents. How much more money does Betty need to buy the wallet?
\\\\
Q1: How much did the house cost?\\
Q2: How much did the repairs increase the value of the house? \\
Q3: What is the new value of the house? \\
Q4: How much profit did he make? 
\\\\
Problem: Betty is saving money for a new wallet which costs \$100. Betty has only half of the money she needs. Her parents decided to give her \$15 for that purpose, and her grandparents twice as much as her parents. How much more money does Betty need to buy the wallet?\\
    \hline
    \textbf{Response}\\
    \hline
    Q1: How much money does Betty have?\\
Q2: How much money did Betty's parents give her?\\
Q3: How much money did Betty's grandparents give her?\\
Q4: How much money does Betty still need to buy the wallet?\\
    \hline
  \end{tabular}
  \caption{Sample ChatGPT input for generating sub-questions and one of the corresponding responses.}
  \label{tab:prompt_qg}
\end{table}

\begin{table}[h]
  \centering
  \begin{tabular}{p{15cm}}
    \hline
    \textbf{Input}\\
    \hline
    You are given the mathematical problems marked with "Problem" and a sequence of questions which should help in solving it. Question number N is marked as "QN:". Based on the problem and sequence of questions answer each of the questions with answer "AN:" and give the answer to the whole problem using "Final answer:" using only the resulting number without adding any additional comments after it.\\\\

Problem: A robe takes 2 bolts of blue fiber and half that much white fiber.  How many bolts in total does it take?\\\\

Q1: How many bolts of white fiber does it take?\\ 
Q2: How many bolts in total does it take?
\\\\
A1: It takes 2/2=1 bolt of white fiber\\
A2: So the total amount of fabric is 2+1=3
\\\\
Final answer: 3
\\\\
Problem: Josh decides to try flipping a house.  He buys a house for \$80,000 and then puts in \$50,000 in repairs.  This increased the value of the house by 150\%.  How much profit did he make?
\\\\
Q1: How much did the house cost?\\
Q2: How much did the repairs increase the value of the house? \\
Q3: What is the new value of the house? \\
Q4: How much profit did he make? 
\\\\
A1: The cost of the house and repairs came out to 80,000+50,000=130,000\\
A2: He increased the value of the house by 80,000*1.5=120,000\\
A3: So the new value of the house is 120,000+80,000=200,000\\
A4: So he made a profit of 200,000-130,000=70,000
\\\\
Final answer: 70000
\\\\
Problem: Betty is saving money for a new wallet which costs \$100. Betty has only half of the money she needs. Her parents decided to give her \$15 for that purpose, and her grandparents twice as much as her parents. How much more money does Betty need to buy the wallet?
\\\\
Q1: How much money does Betty have?\\
Q2: How much money did Betty's parents give her?\\
Q3: How much money did Betty's grandparents give her?\\
Q4: How much money does Betty still need to buy the wallet?\\
    \hline
    \textbf{Response}\\
    \hline
    A1: Betty has 100/2=50 dollars.\\
    A2: Betty's parents gave her 15 dollars.\\
    A3: Betty's grandparents gave her 15*2=30 dollars.\\
    A4: Betty still needs 100-50-15-30=5 dollars.\\\\
    Final answer: 5\\
    \hline
  \end{tabular}
  \caption{Sample ChatGPT input for answering sub-questions and the corresponding response.}
  \label{tab:prompt_qa}
\end{table}

\begin{table}[h]
  \centering
  \begin{tabular}{p{15cm}}
    \hline
    \textbf{Input}\\
    \hline
    You are given the mathematical problem marked with "Problem" and a sequence of sub-questions for solving it. Sub-question number N is marked as "QN:". Based on the problem for each sub-questions decide whether this question is helpful for solving the given problem. An essential property of a good questioning strategy is to ask questions that are directed towards the most critical domain specific content.  Asking the right sequence of relevant questions that can assist in reaching the final goal  is an important part of good questioning. If question repeats any of the previous it is not useful. The question for which answer is given in the problem or can't be answered at all is also not useful. So redundant questions are not good. 
\\\\
For each question output me "QN: $<$Yes/No$>$" and only it where N is the number of the question, e.g. "Q1: $<$Yes/No$>$ Q2: $<$Yes/No$>$ for the first two questions. Do not try to solve the problem anyhow as I'm only interested in the quality of the sub-questions. Strictly follow the output format. Provide answers only for the last given problem.
\\\\
Problem: Janet’s ducks lay 16 eggs per day. She eats three for breakfast every morning and bakes muffins for her friends every day with four. She sells the remainder at the farmers' market daily for \$2 per fresh duck egg. How much in dollars does she make every day at the farmers' market?
\\\\
Q1: How many eggs does Janet sell?\\
Q2: Is duck an animal? \\
Q3: How many eggs does each duck lay? \\
Q4: How much does Janet make at the farmers' market?
\\\\
Q1: Yes\\
Q2: No\\
Q3: No\\
Q4: Yes
\\\\
Problem: A robe takes 2 bolts of blue fiber and half that much white fiber.  How many bolts in total does it take?
\\\\
Q1: How many bolts of white fiber does it take? \\
Q2: How bolts of blue fiber does it take? \\
Q3: How bolts of white fiber does it take? \\
Q4: How many bolts in total does it take? 
\\\\
Q1: Yes\\
Q2: No\\
Q3: No\\
Q4: Yes
\\\\
Problem: Betty is saving money for a new wallet which costs \$100. Betty has only half of the money she needs. Her parents decided to give her \$15 for that purpose, and her grandparents twice as much as her parents. How much more money does Betty need to buy the wallet?
\\\\
Q1: How much money does Betty have?\\
Q2: How much money did Betty's parents give her?\\
Q3: How much money did Betty's grandparents give her?\\
Q4: How much money does Betty still need to buy the wallet?\\
    \hline
    \textbf{Response}\\
    \hline
    Q1: Yes\\
    Q2: Yes\\
    Q3: Yes\\
    Q4: Yes\\
    \hline
  \end{tabular}
  \caption{Sample ChatGPT input for generating sub-questions feedback and one of the corresponding responses.}
  \label{tab:prompt_feedback}
\end{table}

\end{document}

%% file: math_commands.tex
\usepackage{amsmath,amsfonts,bm}

\def\eqref#1{equation~\ref{#1}}
\def\1{\bm{1}}

\DeclareMathAlphabet{\mathsfit}{\encodingdefault}{\sfdefault}{m}{sl}
\SetMathAlphabet{\mathsfit}{bold}{\encodingdefault}{\sfdefault}{bx}{n}

%% file: main.bbl
\begin{thebibliography}{52}
\providecommand{\natexlab}[1]{#1}
\providecommand{\url}[1]{\texttt{#1}}
\expandafter\ifx\csname urlstyle\endcsname\relax
  \providecommand{\doi}[1]{doi: #1}\else
  \providecommand{\doi}{doi: \begingroup \urlstyle{rm}\Url}\fi

\bibitem[Akimov et~al.(2022)Akimov, Kurenkov, Nikulin, Tarasov, and Kolesnikov]{akimov2022let}
Dmitriy Akimov, Vladislav Kurenkov, Alexander Nikulin, Denis Tarasov, and Sergey Kolesnikov.
\newblock Let offline rl flow: Training conservative agents in the latent space of normalizing flows.
\newblock \emph{arXiv preprint arXiv:2211.11096}, 2022.

\bibitem[An et~al.(2021)An, Moon, Kim, and Song]{an2021uncertainty}
Gaon An, Seungyong Moon, Jang-Hyun Kim, and Hyun~Oh Song.
\newblock Uncertainty-based offline reinforcement learning with diversified q-ensemble.
\newblock \emph{Advances in neural information processing systems}, 34:\penalty0 7436--7447, 2021.

\bibitem[Bai et~al.(2022{\natexlab{a}})Bai, Jones, Ndousse, Askell, Chen, DasSarma, Drain, Fort, Ganguli, Henighan, et~al.]{bai2022training}
Yuntao Bai, Andy Jones, Kamal Ndousse, Amanda Askell, Anna Chen, Nova DasSarma, Dawn Drain, Stanislav Fort, Deep Ganguli, Tom Henighan, et~al.
\newblock Training a helpful and harmless assistant with reinforcement learning from human feedback.
\newblock \emph{arXiv preprint arXiv:2204.05862}, 2022{\natexlab{a}}.

\bibitem[Bai et~al.(2022{\natexlab{b}})Bai, Kadavath, Kundu, Askell, Kernion, Jones, Chen, Goldie, Mirhoseini, McKinnon, et~al.]{bai2022constitutional}
Yuntao Bai, Saurav Kadavath, Sandipan Kundu, Amanda Askell, Jackson Kernion, Andy Jones, Anna Chen, Anna Goldie, Azalia Mirhoseini, Cameron McKinnon, et~al.
\newblock Constitutional ai: Harmlessness from ai feedback.
\newblock \emph{arXiv preprint arXiv:2212.08073}, 2022{\natexlab{b}}.

\bibitem[Chen et~al.(2021)Chen, Lu, Rajeswaran, Lee, Grover, Laskin, Abbeel, Srinivas, and Mordatch]{chen2021decision}
Lili Chen, Kevin Lu, Aravind Rajeswaran, Kimin Lee, Aditya Grover, Misha Laskin, Pieter Abbeel, Aravind Srinivas, and Igor Mordatch.
\newblock Decision transformer: Reinforcement learning via sequence modeling.
\newblock \emph{Advances in neural information processing systems}, 34:\penalty0 15084--15097, 2021.

\bibitem[Chen et~al.(2022)Chen, Xu, Gatto, Jain, Kumar, and Chi]{Chen2022OffPolicyAF}
Minmin Chen, Can Xu, Vince Gatto, Devanshu Jain, Aviral Kumar, and Ed~H. Chi.
\newblock Off-policy actor-critic for recommender systems.
\newblock \emph{Proceedings of the 16th ACM Conference on Recommender Systems}, 2022.

\bibitem[Cobbe et~al.(2021)Cobbe, Kosaraju, Bavarian, Chen, Jun, Kaiser, Plappert, Tworek, Hilton, Nakano, et~al.]{cobbe2021training}
Karl Cobbe, Vineet Kosaraju, Mohammad Bavarian, Mark Chen, Heewoo Jun, Lukasz Kaiser, Matthias Plappert, Jerry Tworek, Jacob Hilton, Reiichiro Nakano, et~al.
\newblock Training verifiers to solve math word problems.
\newblock \emph{arXiv preprint arXiv:2110.14168}, 2021.

\bibitem[Diehl et~al.(2021)Diehl, Sievernich, Kr{\"u}ger, Hoffmann, and Bertram]{Diehl2021UMBRELLAUM}
Christopher~P. Diehl, Timo Sievernich, Martin Kr{\"u}ger, Frank Hoffmann, and Torsten Bertram.
\newblock Umbrella: Uncertainty-aware model-based offline reinforcement learning leveraging planning.
\newblock \emph{ArXiv}, abs/2111.11097, 2021.

\bibitem[Fu et~al.(2020)Fu, Kumar, Nachum, Tucker, and Levine]{fu2020d4rl}
Justin Fu, Aviral Kumar, Ofir Nachum, George Tucker, and Sergey Levine.
\newblock D4rl: Datasets for deep data-driven reinforcement learning.
\newblock \emph{arXiv preprint arXiv:2004.07219}, 2020.

\bibitem[Fujimoto \& Gu(2021)Fujimoto and Gu]{fujimoto2021minimalist}
Scott Fujimoto and Shixiang~Shane Gu.
\newblock A minimalist approach to offline reinforcement learning.
\newblock \emph{Advances in neural information processing systems}, 34:\penalty0 20132--20145, 2021.

\bibitem[Ghasemipour et~al.(2022)Ghasemipour, Gu, and Nachum]{ghasemipour2022so}
Kamyar Ghasemipour, Shixiang~Shane Gu, and Ofir Nachum.
\newblock Why so pessimistic? estimating uncertainties for offline rl through ensembles, and why their independence matters.
\newblock \emph{Advances in Neural Information Processing Systems}, 35:\penalty0 18267--18281, 2022.

\bibitem[Gulcehre et~al.(2023)Gulcehre, Paine, Srinivasan, Konyushkova, Weerts, Sharma, Siddhant, Ahern, Wang, Gu, et~al.]{gulcehre2023reinforced}
Caglar Gulcehre, Tom~Le Paine, Srivatsan Srinivasan, Ksenia Konyushkova, Lotte Weerts, Abhishek Sharma, Aditya Siddhant, Alex Ahern, Miaosen Wang, Chenjie Gu, et~al.
\newblock Reinforced self-training (rest) for language modeling.
\newblock \emph{arXiv preprint arXiv:2308.08998}, 2023.

\bibitem[Guss et~al.(2019)Guss, Houghton, Topin, Wang, Codel, Veloso, and Salakhutdinov]{guss2019minerl}
William~H Guss, Brandon Houghton, Nicholay Topin, Phillip Wang, Cayden Codel, Manuela Veloso, and Ruslan Salakhutdinov.
\newblock Minerl: A large-scale dataset of minecraft demonstrations.
\newblock \emph{arXiv preprint arXiv:1907.13440}, 2019.

\bibitem[Hendrycks et~al.(2021)Hendrycks, Burns, Kadavath, Arora, Basart, Tang, Song, and Steinhardt]{hendrycks2021measuring}
Dan Hendrycks, Collin Burns, Saurav Kadavath, Akul Arora, Steven Basart, Eric Tang, Dawn Song, and Jacob Steinhardt.
\newblock Measuring mathematical problem solving with the math dataset.
\newblock \emph{arXiv preprint arXiv:2103.03874}, 2021.

\bibitem[Hinton et~al.(2015)Hinton, Vinyals, and Dean]{hinton2015distilling}
Geoffrey Hinton, Oriol Vinyals, and Jeff Dean.
\newblock Distilling the knowledge in a neural network.
\newblock \emph{arXiv preprint arXiv:1503.02531}, 2015.

\bibitem[Hsieh et~al.(2023)Hsieh, Li, Yeh, Nakhost, Fujii, Ratner, Krishna, Lee, and Pfister]{Hsieh2023DistillingSO}
Cheng-Yu Hsieh, Chun-Liang Li, Chih-Kuan Yeh, Hootan Nakhost, Yasuhisa Fujii, Alexander~J. Ratner, Ranjay Krishna, Chen-Yu Lee, and Tomas Pfister.
\newblock Distilling step-by-step! outperforming larger language models with less training data and smaller model sizes.
\newblock \emph{ArXiv}, abs/2305.02301, 2023.
\newblock URL \url{https://arxiv.org/abs/2305.02301}.

\bibitem[Huang \& Chang(2022)Huang and Chang]{huang2022towards}
Jie Huang and Kevin Chen-Chuan Chang.
\newblock Towards reasoning in large language models: A survey.
\newblock \emph{arXiv preprint arXiv:2212.10403}, 2022.

\bibitem[Jiang et~al.(2023)Jiang, Sablayrolles, Mensch, Bamford, Chaplot, Casas, Bressand, Lengyel, Lample, Saulnier, et~al.]{jiang2023mistral}
Albert~Q Jiang, Alexandre Sablayrolles, Arthur Mensch, Chris Bamford, Devendra~Singh Chaplot, Diego de~las Casas, Florian Bressand, Gianna Lengyel, Guillaume Lample, Lucile Saulnier, et~al.
\newblock Mistral 7b.
\newblock \emph{arXiv preprint arXiv:2310.06825}, 2023.

\bibitem[Juneja et~al.(2023)Juneja, Dutta, Chakrabarti, Manchanda, and Chakraborty]{juneja2023small}
Gurusha Juneja, Subhabrata Dutta, Soumen Chakrabarti, Sunny Manchanda, and Tanmoy Chakraborty.
\newblock Small language models fine-tuned to coordinate larger language models improve complex reasoning.
\newblock \emph{arXiv preprint arXiv:2310.18338}, 2023.

\bibitem[Kojima et~al.(2022)Kojima, Gu, Reid, Matsuo, and Iwasawa]{kojima2022large}
Takeshi Kojima, Shixiang~Shane Gu, Machel Reid, Yutaka Matsuo, and Yusuke Iwasawa.
\newblock Large language models are zero-shot reasoners.
\newblock \emph{Advances in neural information processing systems}, 35:\penalty0 22199--22213, 2022.

\bibitem[Kostrikov et~al.(2021)Kostrikov, Nair, and Levine]{kostrikov2021offline}
Ilya Kostrikov, Ashvin Nair, and Sergey Levine.
\newblock Offline reinforcement learning with implicit q-learning.
\newblock \emph{arXiv preprint arXiv:2110.06169}, 2021.

\bibitem[Kumar et~al.(2020)Kumar, Zhou, Tucker, and Levine]{kumar2020conservative}
Aviral Kumar, Aurick Zhou, George Tucker, and Sergey Levine.
\newblock Conservative q-learning for offline reinforcement learning.
\newblock \emph{Advances in Neural Information Processing Systems}, 33:\penalty0 1179--1191, 2020.

\bibitem[Kumar et~al.(2021)Kumar, Singh, Tian, Finn, and Levine]{kumar2021a}
Aviral Kumar, Anikait Singh, Stephen Tian, Chelsea Finn, and Sergey Levine.
\newblock A workflow for offline model-free robotic reinforcement learning.
\newblock In \emph{5th Annual Conference on Robot Learning}, 2021.
\newblock URL \url{https://openreview.net/forum?id=fy4ZBWxYbIo}.

\bibitem[Kurenkov et~al.(2023)Kurenkov, Nikulin, Tarasov, and Kolesnikov]{kurenkov2023katakomba}
Vladislav Kurenkov, Alexander Nikulin, Denis Tarasov, and Sergey Kolesnikov.
\newblock Katakomba: Tools and benchmarks for data-driven nethack.
\newblock \emph{arXiv preprint arXiv:2306.08772}, 2023.

\bibitem[Lee et~al.(2023)Lee, Phatale, Mansoor, Lu, Mesnard, Bishop, Carbune, and Rastogi]{lee2023rlaif}
Harrison Lee, Samrat Phatale, Hassan Mansoor, Kellie Lu, Thomas Mesnard, Colton Bishop, Victor Carbune, and Abhinav Rastogi.
\newblock Rlaif: Scaling reinforcement learning from human feedback with ai feedback.
\newblock \emph{arXiv preprint arXiv:2309.00267}, 2023.

\bibitem[Levine et~al.(2020)Levine, Kumar, Tucker, and Fu]{levine2020offline}
Sergey Levine, Aviral Kumar, George Tucker, and Justin Fu.
\newblock Offline reinforcement learning: Tutorial, review, and perspectives on open problems.
\newblock \emph{arXiv preprint arXiv:2005.01643}, 2020.

\bibitem[Ling et~al.(2017)Ling, Yogatama, Dyer, and Blunsom]{ling2017program}
Wang Ling, Dani Yogatama, Chris Dyer, and Phil Blunsom.
\newblock Program induction by rationale generation: Learning to solve and explain algebraic word problems.
\newblock \emph{arXiv preprint arXiv:1705.04146}, 2017.

\bibitem[Lyu et~al.(2023)Lyu, Havaldar, Stein, Zhang, Rao, Wong, Apidianaki, and Callison-Burch]{lyu2023faithful}
Qing Lyu, Shreya Havaldar, Adam Stein, Li~Zhang, Delip Rao, Eric Wong, Marianna Apidianaki, and Chris Callison-Burch.
\newblock Faithful chain-of-thought reasoning.
\newblock \emph{arXiv preprint arXiv:2301.13379}, 2023.

\bibitem[Magister et~al.(2023)Magister, Mallinson, Adamek, Malmi, and Severyn]{magister2023teaching}
Lucie~Charlotte Magister, Jonathan Mallinson, Jakub Adamek, Eric Malmi, and Aliaksei Severyn.
\newblock Teaching small language models to reason, 2023.

\bibitem[Nikulin et~al.(2023)Nikulin, Kurenkov, Tarasov, and Kolesnikov]{nikulin2023anti}
Alexander Nikulin, Vladislav Kurenkov, Denis Tarasov, and Sergey Kolesnikov.
\newblock Anti-exploration by random network distillation.
\newblock \emph{arXiv preprint arXiv:2301.13616}, 2023.

\bibitem[OpenAI(2023)]{openai2023gpt}
R~OpenAI.
\newblock Gpt-4 technical report.
\newblock \emph{arXiv}, pp.\  2303--08774, 2023.

\bibitem[Ouyang et~al.(2022)Ouyang, Wu, Jiang, Almeida, Wainwright, Mishkin, Zhang, Agarwal, Slama, Ray, et~al.]{ouyang2022training}
Long Ouyang, Jeffrey Wu, Xu~Jiang, Diogo Almeida, Carroll Wainwright, Pamela Mishkin, Chong Zhang, Sandhini Agarwal, Katarina Slama, Alex Ray, et~al.
\newblock Training language models to follow instructions with human feedback.
\newblock \emph{Advances in Neural Information Processing Systems}, 35:\penalty0 27730--27744, 2022.

\bibitem[Papineni et~al.(2002)Papineni, Roukos, Ward, and Zhu]{papineni2002bleu}
Kishore Papineni, Salim Roukos, Todd Ward, and Wei-Jing Zhu.
\newblock Bleu: a method for automatic evaluation of machine translation.
\newblock In \emph{Proceedings of the 40th annual meeting of the Association for Computational Linguistics}, pp.\  311--318, 2002.

\bibitem[Radford et~al.(2019)Radford, Wu, Child, Luan, Amodei, Sutskever, et~al.]{radford2019language}
Alec Radford, Jeffrey Wu, Rewon Child, David Luan, Dario Amodei, Ilya Sutskever, et~al.
\newblock Language models are unsupervised multitask learners.
\newblock \emph{OpenAI blog}, 1\penalty0 (8):\penalty0 9, 2019.

\bibitem[Sanh et~al.(2019)Sanh, Debut, Chaumond, and Wolf]{sanh2019distilbert}
Victor Sanh, Lysandre Debut, Julien Chaumond, and Thomas Wolf.
\newblock Distilbert, a distilled version of bert: smaller, faster, cheaper and lighter.
\newblock \emph{arXiv preprint arXiv:1910.01108}, 2019.

\bibitem[Shridhar et~al.(2022)Shridhar, Macina, El-Assady, Sinha, Kapur, and Sachan]{shridhar2022automatic}
Kumar Shridhar, Jakub Macina, Mennatallah El-Assady, Tanmay Sinha, Manu Kapur, and Mrinmaya Sachan.
\newblock Automatic generation of socratic subquestions for teaching math word problems.
\newblock \emph{arXiv preprint arXiv:2211.12835}, 2022.

\bibitem[Shridhar et~al.(2023)Shridhar, Stolfo, and Sachan]{shridhar2023distilling}
Kumar Shridhar, Alessandro Stolfo, and Mrinmaya Sachan.
\newblock Distilling reasoning capabilities into smaller language models.
\newblock In \emph{Findings of the Association for Computational Linguistics: ACL 2023}, pp.\  7059--7073, 2023.

\bibitem[Smith et~al.(2022)Smith, Kostrikov, and Levine]{smithWalkParkLearning2022}
Laura Smith, Ilya Kostrikov, and Sergey Levine.
\newblock A {{Walk}} in the {{Park}}: {{Learning}} to {{Walk}} in 20 {{Minutes With Model-Free Reinforcement Learning}}, August 2022.

\bibitem[Snell et~al.(2022)Snell, Kostrikov, Su, Yang, and Levine]{snell2022offline}
Charlie Snell, Ilya Kostrikov, Yi~Su, Mengjiao Yang, and Sergey Levine.
\newblock Offline rl for natural language generation with implicit language q learning.
\newblock \emph{arXiv preprint arXiv:2206.11871}, 2022.

\bibitem[Tarasov et~al.(2022)Tarasov, Nikulin, Akimov, Kurenkov, and Kolesnikov]{tarasov2022corl}
Denis Tarasov, Alexander Nikulin, Dmitry Akimov, Vladislav Kurenkov, and Sergey Kolesnikov.
\newblock Corl: Research-oriented deep offline reinforcement learning library.
\newblock \emph{arXiv preprint arXiv:2210.07105}, 2022.

\bibitem[Tarasov et~al.(2023{\natexlab{a}})Tarasov, Kurenkov, Nikulin, and Kolesnikov]{tarasov2023revisiting}
Denis Tarasov, Vladislav Kurenkov, Alexander Nikulin, and Sergey Kolesnikov.
\newblock Revisiting the minimalist approach to offline reinforcement learning.
\newblock \emph{arXiv preprint arXiv:2305.09836}, 2023{\natexlab{a}}.

\bibitem[Tarasov et~al.(2023{\natexlab{b}})Tarasov, Mbou~Sob, Arbesu, Siboni, Boyer, Skwark, Smit, Bent, and Pretorius]{tarasov2023offline}
Denis Tarasov, Ulrich~Armel Mbou~Sob, Miguel Arbesu, Nima Siboni, Sebastien Boyer, Marcin Skwark, Andries Smit, Oliver Bent, and Arnu Pretorius.
\newblock Offline rl for generative design of protein binders.
\newblock \emph{bioRxiv}, pp.\  2023--11, 2023{\natexlab{b}}.

\bibitem[Team et~al.(2023)Team, Anil, Borgeaud, Wu, Alayrac, Yu, Soricut, Schalkwyk, Dai, Hauth, et~al.]{team2023gemini}
Gemini Team, Rohan Anil, Sebastian Borgeaud, Yonghui Wu, Jean-Baptiste Alayrac, Jiahui Yu, Radu Soricut, Johan Schalkwyk, Andrew~M Dai, Anja Hauth, et~al.
\newblock Gemini: a family of highly capable multimodal models.
\newblock \emph{arXiv preprint arXiv:2312.11805}, 2023.

\bibitem[Touvron et~al.(2023)Touvron, Lavril, Izacard, Martinet, Lachaux, Lacroix, Rozi{\`e}re, Goyal, Hambro, Azhar, et~al.]{touvron2023llama}
Hugo Touvron, Thibaut Lavril, Gautier Izacard, Xavier Martinet, Marie-Anne Lachaux, Timoth{\'e}e Lacroix, Baptiste Rozi{\`e}re, Naman Goyal, Eric Hambro, Faisal Azhar, et~al.
\newblock Llama: Open and efficient foundation language models.
\newblock \emph{arXiv preprint arXiv:2302.13971}, 2023.

\bibitem[Tunstall et~al.(2023)Tunstall, Beeching, Lambert, Rajani, Rasul, Belkada, Huang, von Werra, Fourrier, Habib, et~al.]{tunstall2023zephyr}
Lewis Tunstall, Edward Beeching, Nathan Lambert, Nazneen Rajani, Kashif Rasul, Younes Belkada, Shengyi Huang, Leandro von Werra, Cl{\'e}mentine Fourrier, Nathan Habib, et~al.
\newblock Zephyr: Direct distillation of lm alignment.
\newblock \emph{arXiv preprint arXiv:2310.16944}, 2023.

\bibitem[Vaswani et~al.(2017)Vaswani, Shazeer, Parmar, Uszkoreit, Jones, Gomez, Kaiser, and Polosukhin]{vaswani2017attention}
Ashish Vaswani, Noam Shazeer, Niki Parmar, Jakob Uszkoreit, Llion Jones, Aidan~N Gomez, {\L}ukasz Kaiser, and Illia Polosukhin.
\newblock Attention is all you need.
\newblock \emph{Advances in neural information processing systems}, 30, 2017.

\bibitem[Vinyals et~al.(2019)Vinyals, Babuschkin, Czarnecki, Mathieu, Dudzik, Chung, Choi, Powell, Ewalds, Georgiev, et~al.]{vinyals2019grandmaster}
Oriol Vinyals, Igor Babuschkin, Wojciech~M Czarnecki, Micha{\"e}l Mathieu, Andrew Dudzik, Junyoung Chung, David~H Choi, Richard Powell, Timo Ewalds, Petko Georgiev, et~al.
\newblock Grandmaster level in starcraft ii using multi-agent reinforcement learning.
\newblock \emph{Nature}, 575\penalty0 (7782):\penalty0 350--354, 2019.

\bibitem[Wang et~al.(2022)Wang, Wei, Schuurmans, Le, Chi, Narang, Chowdhery, and Zhou]{wang2022self}
Xuezhi Wang, Jason Wei, Dale Schuurmans, Quoc Le, Ed~Chi, Sharan Narang, Aakanksha Chowdhery, and Denny Zhou.
\newblock Self-consistency improves chain of thought reasoning in language models.
\newblock \emph{arXiv preprint arXiv:2203.11171}, 2022.

\bibitem[Wei et~al.(2022)Wei, Wang, Schuurmans, Bosma, Xia, Chi, Le, Zhou, et~al.]{wei2022chain}
Jason Wei, Xuezhi Wang, Dale Schuurmans, Maarten Bosma, Fei Xia, Ed~Chi, Quoc~V Le, Denny Zhou, et~al.
\newblock Chain-of-thought prompting elicits reasoning in large language models.
\newblock \emph{Advances in Neural Information Processing Systems}, 35:\penalty0 24824--24837, 2022.

\bibitem[Yang et~al.(2022)Yang, Bai, Ma, Wang, Zhang, and Han]{yang2022rorl}
Rui Yang, Chenjia Bai, Xiaoteng Ma, Zhaoran Wang, Chongjie Zhang, and Lei Han.
\newblock Rorl: Robust offline reinforcement learning via conservative smoothing.
\newblock \emph{Advances in Neural Information Processing Systems}, 35:\penalty0 23851--23866, 2022.

\bibitem[Yuan et~al.(2023)Yuan, Yuan, Li, Dong, Tan, and Zhou]{Yuan2023ScalingRO}
Zheng Yuan, Hongyi Yuan, Cheng Li, Guanting Dong, Chuanqi Tan, and Chang Zhou.
\newblock Scaling relationship on learning mathematical reasoning with large language models.
\newblock \emph{ArXiv}, abs/2308.01825, 2023.
\newblock URL \url{https://arxiv.org/abs/2308.01825}.

\bibitem[Zhou et~al.(2022)Zhou, Sch{\"a}rli, Hou, Wei, Scales, Wang, Schuurmans, Cui, Bousquet, Le, et~al.]{zhou2022least}
Denny Zhou, Nathanael Sch{\"a}rli, Le~Hou, Jason Wei, Nathan Scales, Xuezhi Wang, Dale Schuurmans, Claire Cui, Olivier Bousquet, Quoc Le, et~al.
\newblock Least-to-most prompting enables complex reasoning in large language models.
\newblock \emph{arXiv preprint arXiv:2205.10625}, 2022.

\end{thebibliography}
